Original Article

**Title:** Automated classification of stems and leaves of potted plants based on point cloud data

**Author:** Zichu Liu[1], Qing Zhang[1], Pei Wang[1*], Zhen Li[1], Huiru Wang[1]

**Address:** [1]College of Science, Beijing Forestry University，No.35 Qinghua East Road, Haidian District, Beijing 100083, China

**Corresponding author:** Pei Wang[1*]

[*]**For correspondence. E-mail:** wangpei@bjfu.edu.cn

**ABSTRACT**

The accurate classification of plant organs is a key step in monitoring the growing status and physiology of plants. A classification method was proposed to classify the leaves and stems of potted plants automatically based on the point cloud data of the plants, which is a nondestructive acquisition. The leaf point training samples were automatically extracted by using the three-dimensional convex hull algorithm, while stem point training samples were extracted by using the point density of a two-dimensional projection. The two training sets were used to classify all the points into leaf points and stem points by utilizing the support vector machine (SVM) algorithm. The proposed method was tested by using the point cloud data of three potted plants and compared with two other methods, which showed that the proposed method can classify leaf and stem points accurately and efficiently.

**Key words:** point cloud data, automated classification, SVM, leaves samples, stem samples

# 1. Introduction

Plant organs and their characteristics are very important to many plant studies. They can be used to monitor plant growth status and study plant physiological characteristics. Williams and Ayars (2005) measured grapevine organs to estimate plant photosynthesis. Conde (2011) estimated the effects of different nutrients on plant growth through measuring the growth statuses of different organs. Davi et al. (2005) studied carbon-water circulation through tree canopy structures. Plant organ research also plays an important role in environmental governance. Huang Huimin et al. (2019) studied Broussonetia papyrifera organs to study the effects of saline-alkali stress on plant morphology and growth. Fitter (1987) studied the adaptability of plants in different environments by comparing different root structures. Sampling plant organs was the most common method in practice in the above studies. However, the traditional destructive sampling method usually requires cutting the plant organs for measurement, which is time-consuming and destructive. Traditional methods not only require a large amount of time and energy but also result in the destruction and deforestation of plants and trees, which will have negative consequences on the ecological environment when conducting a large number of tests and data collections.

3D laser scanning technology can obtain the point cloud data of plants accurately and quickly, which provides a solution for nondestructive data collection and the fine-grained analyses of plant organs. The point cloud data of plants, collected with high precision and high density, record the plant geometry accurately, which is good for the analyses of plant organs. This technology has been applied in related studies on plants and agroforestry (Zheng and Moskal, 2012; Caccamo et al., 2018; Morsdorf et al., 2007; Maltamo et al., 2010), and the use of plant point cloud data to accurately classify and identify different organs of plants is a prerequisite for conducting the above studies. Many scholars and studies have proposed related classification methods. Wahabzada et al. (2015) proposed a data-driven method for plant organ segmentation when plants are occluded. Yun et al. (2013) proposed a classification method for constructing a covariance matrix based on neighborhood information. The method extracts the feature vector of each point from the scanned point cloud data and then uses the manifold learning method to reduce the point cloud. Paulus et al. (2013) used plant histological surface histograms to classify individual plant organs through a fully automated system and applied the proposed method to wheat estimation. Hétroy-Wheeler et al. (2016) proposed a semi-automatic point cloud classification method to divide small tree seedlings into leaves, petioles, and stems. The method has strong robustness, and its false-positive rate and false-negative rate were both approximately 1%. Frasson et al. (2010) proposed a method for using a preprocessed grid to represent plant organs and manually dividing the meshes into different morphological regions. Ma et al. (2016) used a geometric feature-based automatic forest point classification (GAFPC) algorithm to divide tree point cloud data into photosynthetic, non-photosynthetic forest canopy components, and bare earth. Ferrara et al. (2018) proposed a method for using voxel density and density-based spatial clustering of applications with noise (DBSCAN) algorithm to divide cork oak trees into wood points and non-wood points. Tao et al. (2015) proposed a geometric method for leaves and wood classification based on different shapes of trunk/branch boundaries and leaf clusters. Dey et al. (2012) used the SVM method to classify the organs of grapevines.

In general, for rough manual sampling, the results are subjectively influenced by the operator and the sample. For careful artificial sampling classification, it is necessary to make repeated sample selection attempts according to the different morphological characteristics and physiological structures of plants. To obtain better classification results, the complicated operation caused by careful artificial sampling consumes considerable time and effort, which increases the research cycle and cost. Therefore, automation in the process of selecting sample points and classification can greatly reduce the time cost of related work and has great significance in practical applications.

In this paper, we propose an automated SVM classification method based on the spatial distribution characteristics and density distribution characteristics of plant point cloud data, experiment on three plant point cloud data sets, and discuss the feasibility, advantages and classification accuracy of this algorithm.

The steps of this work are as follows: (1) The point cloud data of three different potted plants are scanned and processed. (2) Next, an automated classification method is proposed. (3) The processed point cloud data are then classified. (4) The standard classification result is subsequently constructed. (5) The proposed method is then compared with two other methods. (6) Finally, the compared results are discussed, and the conclusions are drawn.

## 2. Materials and Methods

*2.1 Experimental data*

In the experiments, three potted plants were scanned by using the HDI 120A-B scanner (manufactured by the 3D LMI technologies company), which has the ability to acquire high-precision point cloud data with structured light. This device is composed of an imaging module and a light source module, and its overall size is small, which greatly reduces the requirements for the scanner operator, greatly improves the scanning efficiency and maintains a high scanning accuracy. The specific information of this 3D scanner is shown in Table 1.

| 3D SCANNER | LMI HDI 120 |
|---|---|
| Camera | $2 \times 13000$ pixel |
| Scanning Software | FlexScan3D |
| Scan Speed | 0.3 second per scan |
| Field of View | $124 \times 120\ mm - 192 \times 175\ mm$ |
| Average Number of Points | 985000 per scan |
| Average Number of Polygons | 19700 per scan |
| Point-to-Point Distance | 0.162 mm |
| Accuracy | $\pm 0.02\ mm$ |

TABLE 1. *Information of HDI 120A-B 3D scanner.*

Three different potted plants were scanned in the experiments: Zamioculas zamiifolia, Pachyphytum bracteosum, and Dieffenbachia picta. As shown in Fig. 1, the left plant is Zamioculas zamiifolia, the middle plant is Pachyphytum bracteosum and the right plant is Dieffenbachia picta.

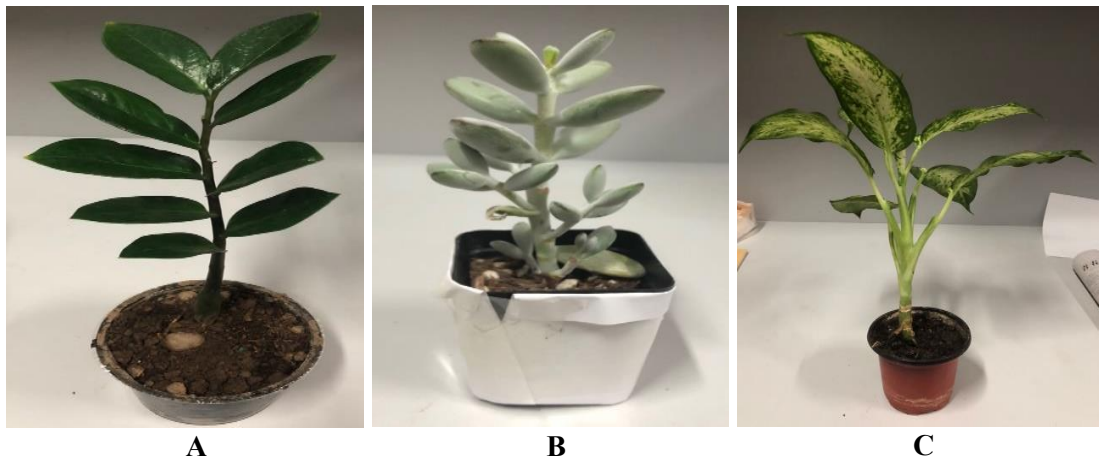

FIG. 1. Three scanned plant pictures.
A: Zamioculas zamiifolia. B: Pachyphytum bracteosum. C: Dieffenbachia picta.

To obtain more visual test results and better classification results, the raw point cloud data of the three plants were denoised and rotated to make the stems of the three plants as perpendicular as possible relative to the XOY plane of the three-dimensional coordinate system, as shown in Fig. 2.

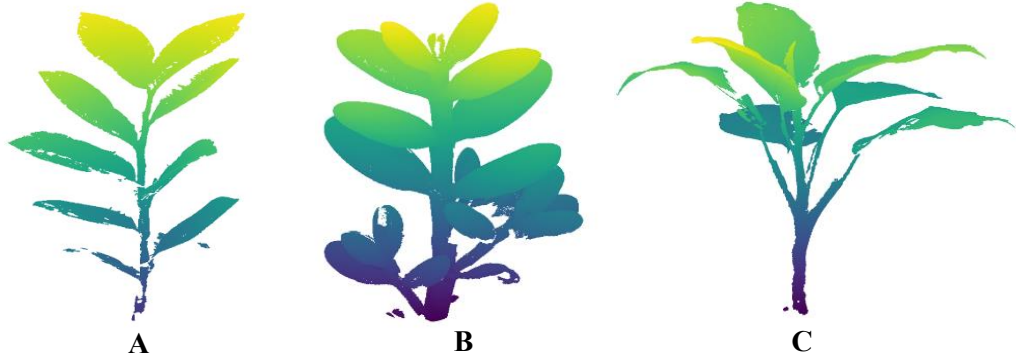

FIG. 2. Point cloud data of the three plants.
A: Zamioculas zamiifolia. B: Pachyphytum bracteosum. C: Dieffenbachia picta.

The numbers of points of these three potted plants after denoising and the information of the circumscribed cuboids of the three plants are listed in Table 2.

| PLANT | X length/mm | Y length/mm | Z length/mm | Number |
| --- | --- | --- | --- | --- |
| Zamioculas zamiifolia | 137.7650 | 108.8128 | 206.4078 | 1044220 |
| Pachyphytum bracteosum | 73.4398 | 72.7338 | 104.5292 | 871577 |
| Dieffenbachia picta | 248.8156 | 224.9988 | 269.1463 | 2691531 |

TABLE 2. *Point-related information of the three plants.*

*2.2 Method*

In this paper, an automated classification method was proposed to classify the leaves and stems of the potted plants. Two different methods were proposed to automatically acquire the training sets of the leaves and stems.

Considering the spatial structure of the potted plants, that is, the leaf point clouds of the plants were distributed on the periphery of the point cloud data, the training sets of the leaf samples were selected by using the three-dimensional convex hull algorithm. First, we constructed a 3D convex hull from the point cloud data and then selected leaf sample points based on their turning points.

Because the stem point cloud data are mainly concentrated in the middle part of the plant point cloud data, it is difficult to directly obtain its sample point distribution. Therefore, we used the random sampling and grid projection density methods to automatically obtain the sample points of the stem, and the training sets of the stems were selected.

Then, the classification results were obtained by using the SVM algorithm with the training sets. The process of the proposed algorithm is shown in Fig. 3.

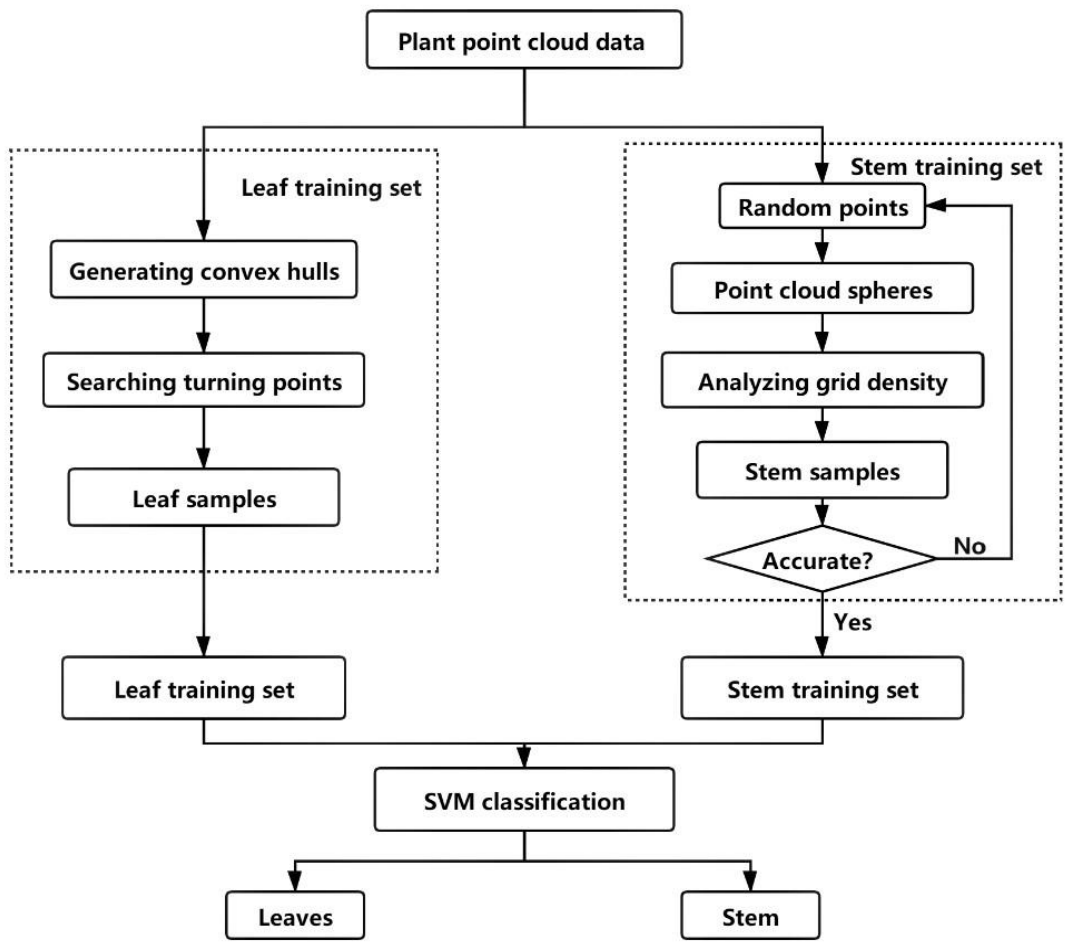

FIG. 3.　Flowchart of the method.

The specific process is as follows: First, a three-dimensional convex hull algorithm was used to obtain the 3D convex hull of the data, and the turning point set $L = \{(x_i, y_i, z_i) | i = 1, \cdots m\}$ was taken as the leaf sample point set, where $(x_i, y_i, z_i)$ is the turning point and $m$ is the number of turning points. Second, each point $(x_i, y_i, z_i)$ was chosen as the center, $r_1$ was set as the radius, leaf sample point spheres were made, and the points inside each sphere were selected as the leaf training set:

$$X_{Leaf} = \{(x,y,z)|(x-x_i)^2 + (y-y_i)^2 + (z-z_i)^2 < r_1^2, (x,y,z) \in D, i = 1,\cdots m\},$$

where $D$ denotes the plant point cloud data and $r_1$ denotes the selection radius of the leaf training set. Third, $p$ points of the point cloud data were randomly sampled, the grid density of each point was calculated, the grid density of each point was set as $m_a$, where $a = 1,\cdots,p$, and then $n$ points with the highest grid density were chosen. If the points chosen at the last step have a good distribution and were the right stem points, these points were regarded as the stem sample point set $S = \{(x_j, y_j, z_j)|j = 1,\cdots n\}$; if not, the stem sample points were reselected until the chosen points were correct. Then, each point $(x_j, y_j, z_j)$ was chosen as the center, $r_2$ was set as the radius, stem sample point spheres were created and the points inside each sphere were selected as the stem training set:

$$X_{Stem} = \{(x,y,z)|(x-x_j)^2 + (y-y_j)^2 + (z-z_j)^2 < r_2^2, (x,y,z) \in D, j = 1,\cdots n\},$$

where $data$ denotes the plant point cloud data and $r_2$ denotes the selection radius of the stem training set. Finally, we marked set $X_{Leaf}$ and set $X_{Stem}$ as class1 and class2, input them into the SVM classifier, classified the point cloud data and analyzed the results.

*2.3 Constructing the training sets of leaves and stems*

*2.3.1 Constructing the leaf training sets*

In this section, the 3D convex hull algorithm was used to achieve the automatic selection of leaf sample points.

A convex hull is a concept under the computational geometry branch in mathematics. The convex hull of a given set $X$ refers to the intersection of all convex sets containing the set $X$ in the real vector space $V$. As shown in Fig. 4, a three-dimensional point set including 16 points was set as $P$. The smallest convex polyhedron shown in the figure is the three-dimensional convex hull of set $P$. The green points are the turning points of this convex hull, and the red points are inside the convex hull.

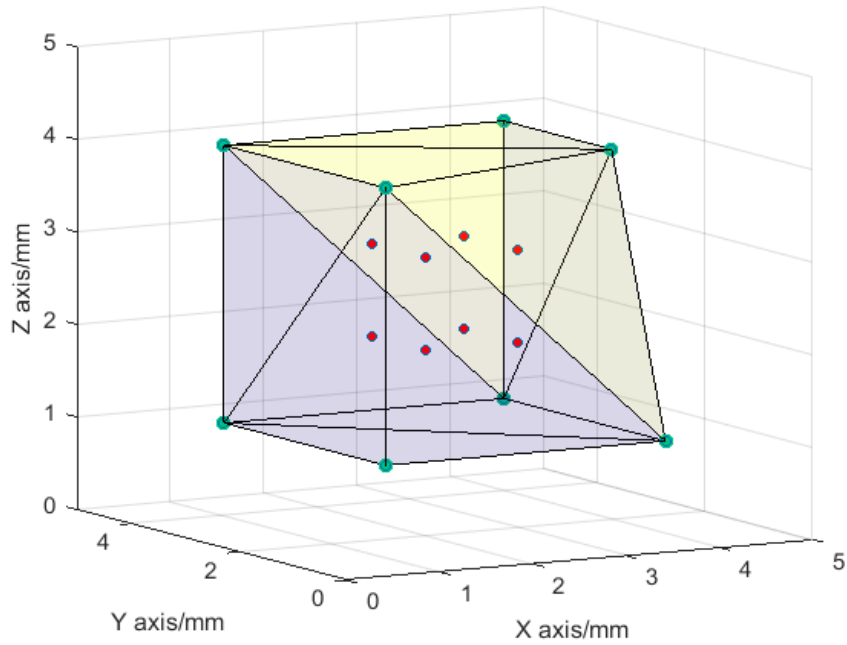

FIG. 4. Illustration of a three-dimensional convex hull.

In this paper, the smallest convex polyhedron containing all point cloud data can be obtained by obtaining the three-dimensional convex hull. Since the stem is generally inside the plant and the points of the leaf tips are generally in the outermost part of the plant, the turning points of the obtained three-dimensional convex hull, that is, the apexes of the smallest convex polyhedron, are usually the points of the leaf tips; thus, these turning points were selected as sample points for the plant leaves. Then, we chose each leaf sample point as the center, set $r_1$ as the radius, made spheres and selected the points inside the spheres as the leaf training set.

*2.3.2 Constructing the stem training sets*

A stem sample point selection method based on the grid densities was proposed. First, n points were selected randomly from the plant point cloud data. Second, the projection grid density of each point was calculated. Then, the few points with the highest density were chosen as the stem sample points. If the selected stem sample points were accurate and well distributed, the stem training set could be obtained by these sample points. If the sample point selection result was not accurate, the parameters were re-entered according to the specific morphology of the plant to select the sample points again.

In the experiment, first, $n$ points of the plant point cloud data were sampled randomly. Second, $p_i$ was selected as the center, $r$ was set as the radius, and the sphere $s_i$ was created, where $p_i$ denotes each randomly sampled point, $i = 1, \cdots, n$. Then, all the points inside the sphere $s_i$ were projected onto the XOY plane to obtain the 2D projection. The circum-square of the circle on the XOZ plane projected by the sphere $s_i$ was then made, and this square was meshed. Next, the number $num_i$ of grids occupied by the projections of all points inside the sphere $s_i$ were counted, with the grid density $m_i$ of each point $p_i$ given as:

$$m_i = \frac{l}{num_i},$$

where $l$ is the number of points in the sphere $s_i$.

Then, some points with the highest grid density were selected as the stem sample points. Finally, for each stem point, $r_2$ was set as the radius, a sphere was created and all the points inside the spheres were regarded as the stem training set.

Take Zamioculas zamiifolia as an example; a point $a_1$ on the stem and a point $a_2$ on the leaf were selected to demonstrate the above process, as shown in Fig. 5. The centers of the two spheres are $a_1$ and $a_2$, and the radii of the two spheres are both 0.5 mm.

In the experiment, the points in the spheres were projected onto the XOY plane with a 0.1 mm grid spacing. As highlighted in Fig. 5, the two projections were obviously different in terms of the point distribution. In the stem sphere, there were 52 points that were projected into 10 grids. However, in the leaf sphere, there were 99 points that were projected into 43 grids. The points per grid of the two projections were 5.2 and 2.3023, respectively, which had a significant difference that was used to discriminate the stem and the leaves in our experiment.

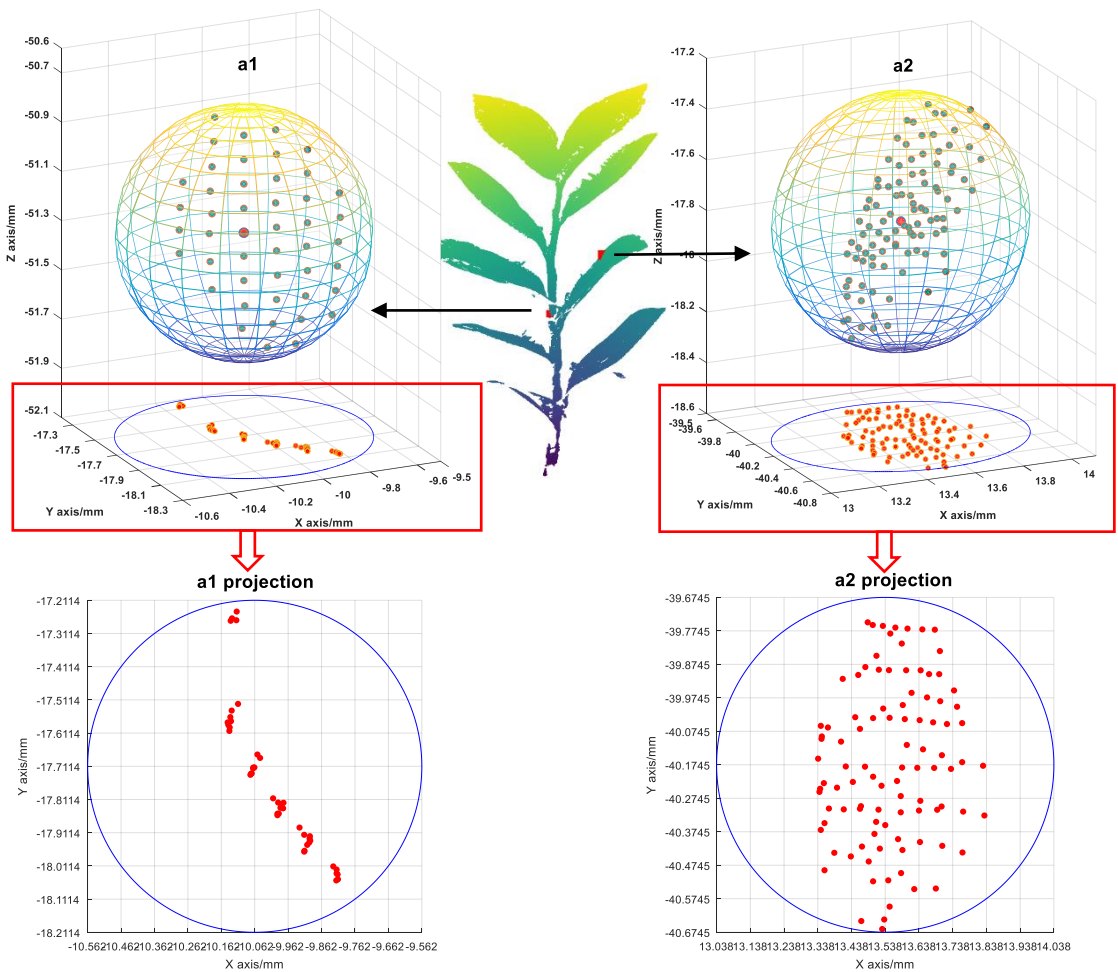

FIG. 5. Sphere projections on the XOY plane.
Left: Stem point sphere a1 projection.    Right: Leaf point sphere a2 projection.

*2.4 Classification by using the SVM method*

The SVM method can separate all data samples by selecting a hyperplane in space. The distance from all original data in the sample set to the hyperplane is the shortest. The plant point cloud data are three-dimensional data, so the classification of the plant point cloud data is nonlinear.

For the nonlinear classification, the samples can be mapped from the original space to another feature space of higher dimensions to finally make the samples present linear separability in the high-dimensional space. Then, the kernel function $K(x_i, x_j)$ can be used to solve the nonlinear problem (Zhang, 2000). Commonly used kernel functions are the linear kernel, polynomial kernel and radial basis function (RBF) kernel.

In this paper, the training sets of the stem and leaves were used to classify the plant point cloud data into stem points and leaf points by using the SVM method. Because the potted plant point cloud data are three-dimensional, a proper kernel function should be selected for classification. After conducting several experiments, the RBF kernel was finally selected for its better classification results. The RBF kernel function can map a sample to an infinite-dimensional space for better classification results and has fewer numerical difficulties. In the previous experiments, the RBF kernel had less computational complexity in MATLAB and occupied less computer memory compared to other kernel functions.

## 3. Results

In this section, the selected samples and the results of the proposed method are demonstrated. Meanwhile, three different methods are used to classify the plant point cloud data for comparison and evaluation.

*3.1 Constructing the standard result of classification*

In this section, to evaluate the different methods, an accurate classification result was obtained by manual sampling and regarded as a comparison criterion. The manually selected leaf points were set as standard leaf classification points, and the manually selected stem points were set as standard stem classification points.

The standard classification results are shown in the first column of Fig. 12A, in which the left plant is Zamioculas zamiifolia, the middle plant is Pachyphytum bracteosum and the right plant is Dieffenbachia picta. The numbers of standard classification result points are listed in Table 5.

*3.2 Two classification methods for comparison*

In this section, to evaluate the proposed method, two different classification methods are given for comparison.

*3.2.1 Random selection classification*

First, a certain number of points in the data were randomly selected. Second, the points at the leaf were regarded as the leaf sample point set $L$, and the same parameter $r_1$ in the above was set as the radius to establish the leaf training point set $X_{Leaf}$. Meanwhile, the points at the stem were regarded as the stem sample point set $S$, and the same parameter $r_2$ in the above was set as the radius to establish the stem training point set $X_{Stem}$. Then, the leaf and stem training sets $X_{Leaf}$ and $X_{Stem}$ were marked as class1 and class2, respectively, and were input into the SVM classifier. Finally, the RBF kernel function was selected for classification.

The random selection classification results are shown in Fig. 12B, in which the plants are Zamioculas zamiifolia, Pachyphytum bracteosum and Dieffenbachia picta (from left to right).

*3.2.2 Artificial selection classification*

In this part, after repeated selection and classification experiments, the sample points with better results were finally selected. Although the most accurate results were obtained, they required considerable time. First, some points at the tips of leaves were selected artificially as the leaf tip point set $L_1$, and some points in the roots of leaves were selected as the leaf root point set $L_2$. Second, $L_1 \cup L_2$ was set as the leaf sample point set $L$. Then, the same parameter $r_1$ in the above was taken as the radius to establish the leaf training point set $X_{Leaf}$. Next, some stem points with a uniform distribution were selected as the stem sample point set $S$, and the same parameter $r_2$ in the above was set as the radius to establish the stem training point set $X_{Stem}$. Then, the leaf and stem training sets were marked as class1 and class2, respectively, and were input into the SVM classifier. Finally, the RBF kernel function was selected for classification.

The artificial selection classification results are shown in Fig. 12C, in which the plants are Zamioculas zamiifolia, Pachyphytum bracteosum and Dieffenbachia picta (from left to right).

**3.3** *Comparing the results of the methods*

In this section, to evaluate the automated classification method and draw conclusions, the sample point selection results and classification results of the three methods are compared.

*3.3.1 Leaf sample point selection results*

By using the proposed method, the 3D convex hulls of the three plants were calculated based on the point cloud data, and the turning points of these convex hulls were regarded as leaf sample points. The numbers of leaf sample points of Zamioculas zamiifolia, Pachyphytum bracteosum and Dieffenbachia picta were 110, 2373, and 206, respectively. The selected leaf samples are shown in Fig. 6. Centering on the leaf sample points, the leaf sample sets were constructed with a radius of 0.2 mm. The numbers of sample points and the numbers of points in the sample training sets are listed in Table 3.

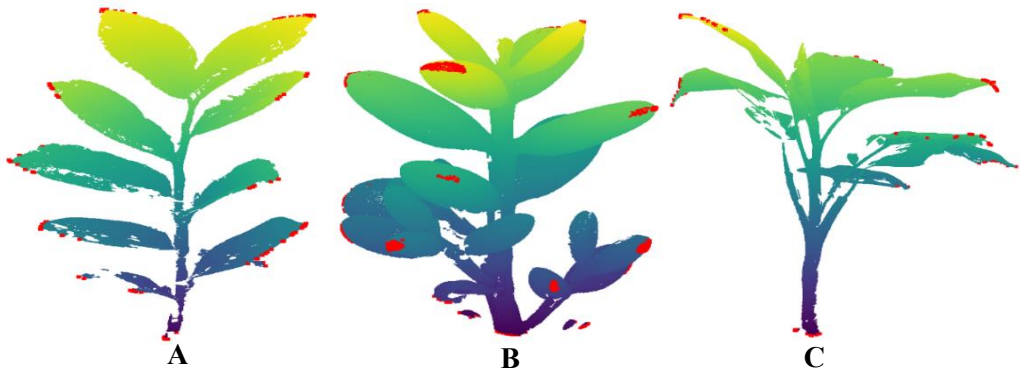

FIG. 6. Leaf sample points of automated selection method.
A: Zamioculas zamiifolia. B: Pachyphytum bracteosum. C: Dieffenbachia picta.

| Plant | Number of leaf sample points | $r_1$ | Number of sample training set points |
|---|---|---|---|
| Zamioculas zamiifolia | 110 | $0.2\ mm$ | 544 |
| Pachyphytum bracteosum | 2373 | $0.2\ mm$ | 21860 |
| Dieffenbachia picta | 206 | $0.2\ mm$ | 1137 |

TABLE 3. *Leaf sample points and leaf sample training sets of the three plants.*

The leaf sample points of the random selection method are shown in Fig. 7, and the leaf sample points of the artificial selection method are shown in Fig. 8.

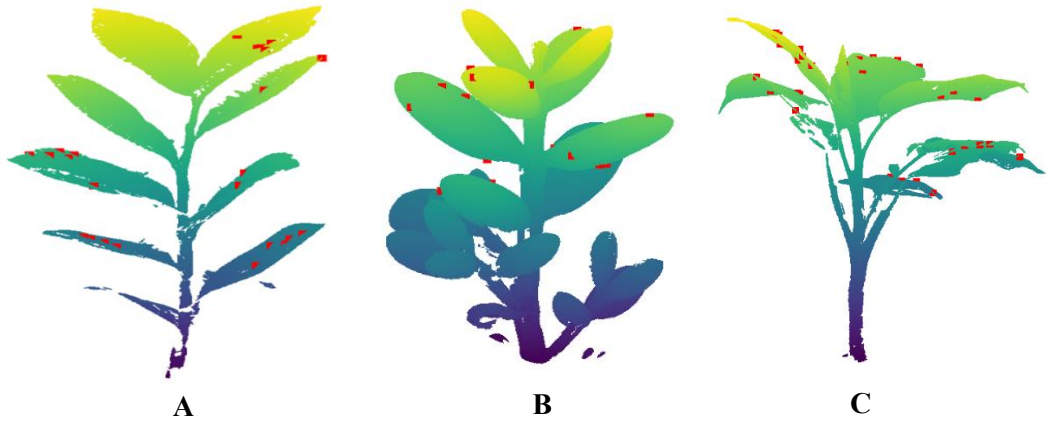

FIG. 7. Leaf sample points of random selection method.
A: Zamioculas zamiifolia. B: Pachyphytum bracteosum. C: Dieffenbachia picta.

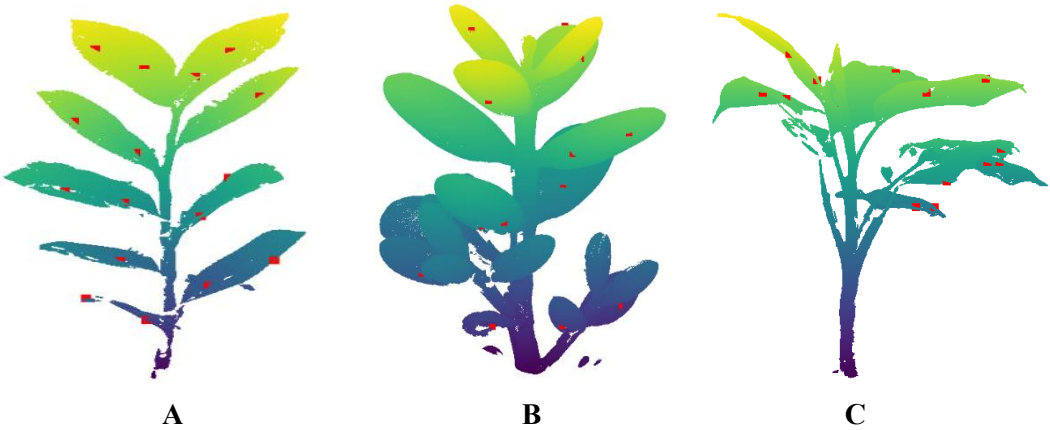

FIG. 8. Leaf sample points of artificial selection method.
A: Zamioculas zamiifolia. B: Pachyphytum bracteosum. C: Dieffenbachia picta.

*3.3.2 Stem sample point selection results*

When selecting the stem sample points by using the proposed method, according to the different plant characteristics, we should choose different $r$ values. The different results obtained by the different values are given in Fig. 9, in which A is Zamioculas zamiifolia, B is Pachyphytum bracteosum and C is Dieffenbachia picta.

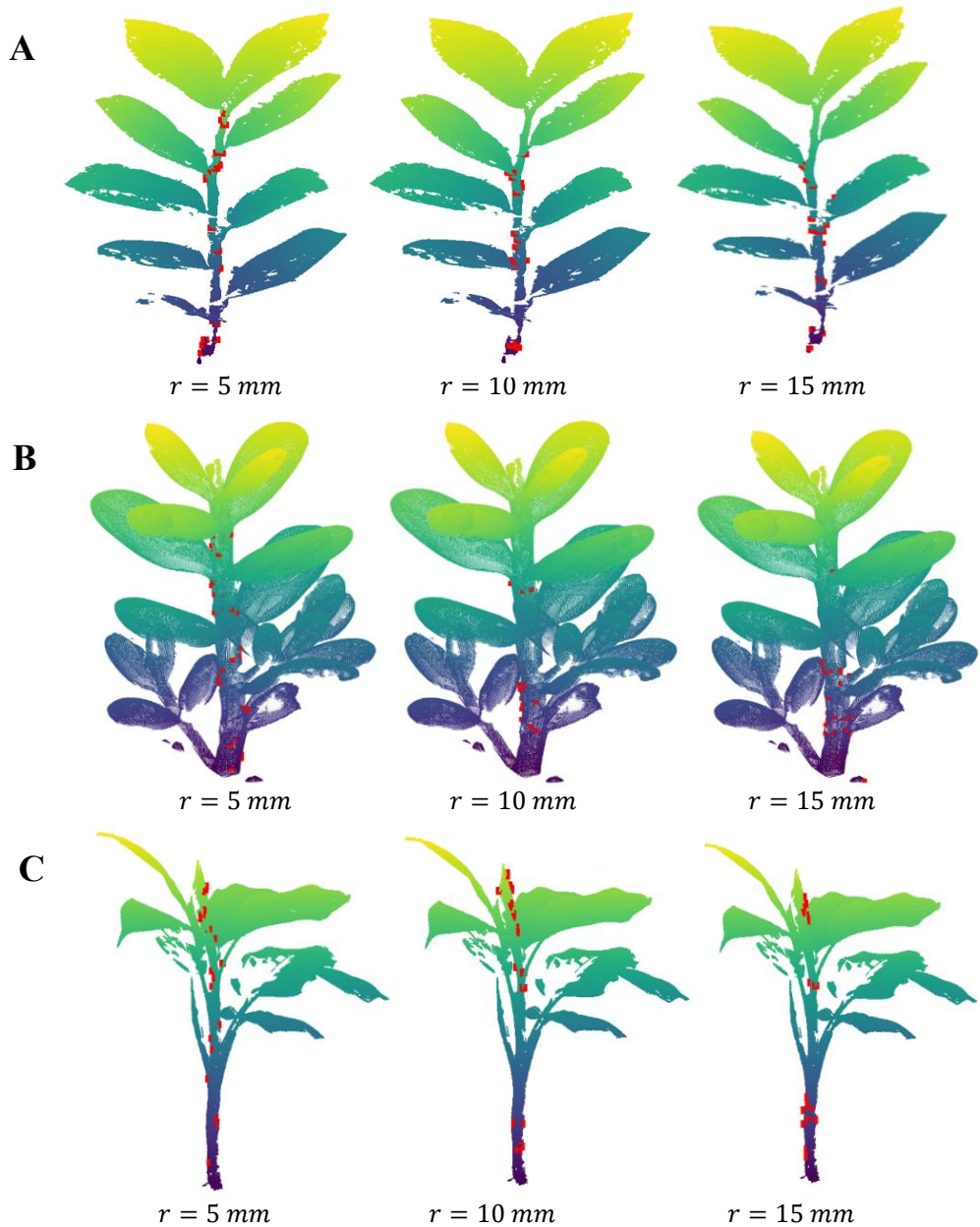

FIG. 9. Stem sample points selected by using different values of $r$.
A: Zamioculas zamiifolia. B: Pachyphytum bracteosum. C: Dieffenbachia picta.

Because these three potted plants are small, their stems are thin; thus, we chose 500 points randomly, with the parameter $r = 5\ mm$ for selecting the stem sample points and $r_2 = 0.2\ mm$ for establishing the stem training point set. Then, 20, 30 and 20 points with the highest grid density were selected as stem sample points of Zamioculas zamiifolia, Pachyphytum bracteosum and Dieffenbachia picta, respectively; among them, Pachyphytum bracteosum has a thick stem, requiring more points to be selected to cover the whole stem. Finally, the numbers of stem training points were 201, 280, and 143 for Zamioculas zamiifolia, Pachyphytum bracteosum and Dieffenbachia picta, respectively, as listed in Table 4.

Zamioculas zamiifolia has a straight and thin stem and wide leaves; thus, the result of the stem sample points obtained by the grid density method was the best. Dieffenbachia picta has a thin and short stem, and some leaves are shaped like stems; thus, this plant has a worse result, which we examine in the discussion section. The specific stem sample point numbers, i.e., $r$ and $r_2$, are also listed in Table 4.

| Plant | Stem sample points | $r$ | $r_2$ | Sample training set points |
|---|---|---|---|---|
| Zamioculas zamiifolia | 20 | $5\ mm$ | $0.2\ mm$ | 201 |
| Pachyphytum bracteosum | 30 | $5\ mm$ | $0.2\ mm$ | 280 |
| Dieffenbachia picta | 20 | $5\ mm$ | $0.2\ mm$ | 143 |

TABLE 4. *Stem sample points and stem sample training sets of the three plants.*

The random selection stem sample points are shown in Fig. 10, and the artificial selection stem sample points are shown in Fig. 11.

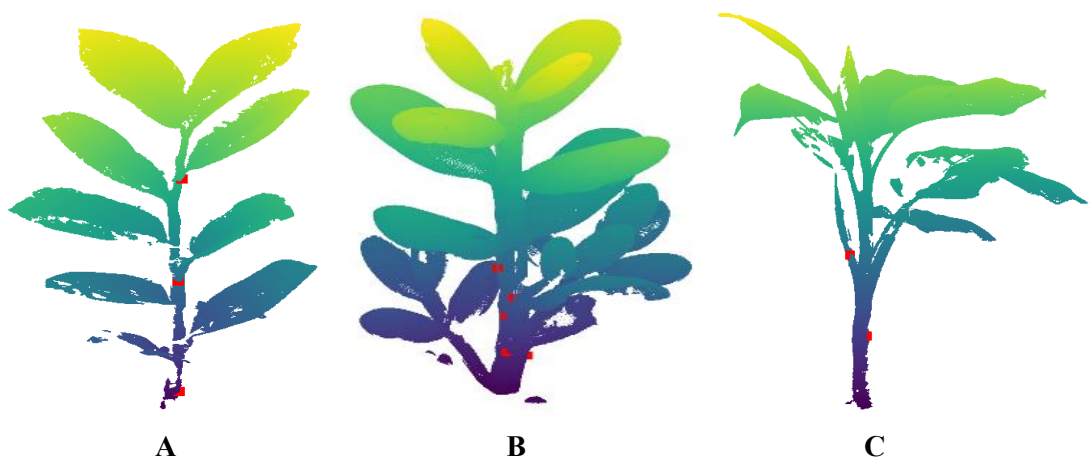

FIG. 10. Stem sample points of random selection method.
A: Zamioculas zamiifolia. B: Pachyphytum bracteosum. C: Dieffenbachia picta.

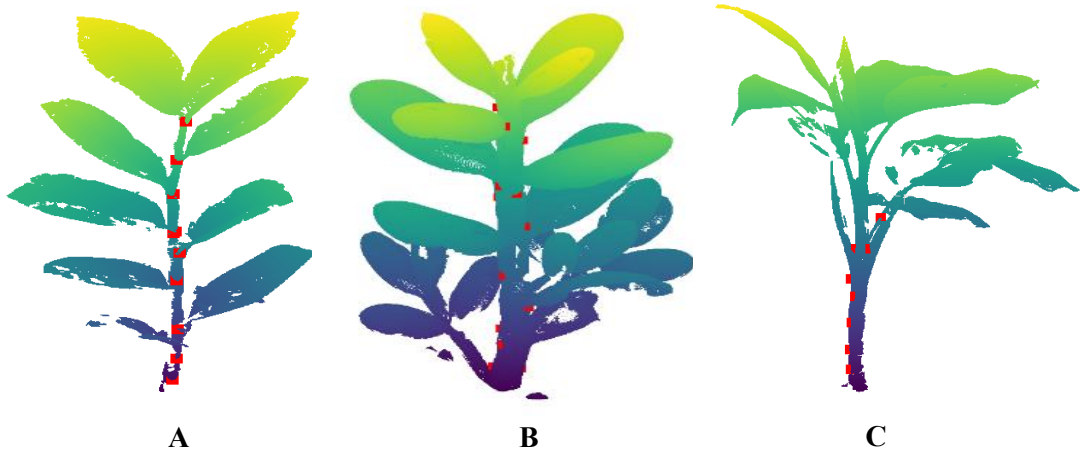

FIG. 11. Stem sample points of artificial selection method.
A: Zamioculas zamiifolia. B: Pachyphytum bracteosum. C: Dieffenbachia picta.

*3.3.3 Classification results*

After selecting the stem training point set and leaf training point set, the SVM algorithm and RBF kernel function were used to classify the three plants. The automated classification results are shown in Fig. 12D, in which the plants are Zamioculas zamiifolia, Pachyphytum bracteosum and Dieffenbachia picta (from left to right).

As shown in Fig. 12, the results of three different methods (Fig. 12B, C, D) were compared to the standard results (Fig. 12A), and as we can see in this figure, the automated classification results of Zamioculas zamiifolia and Pachyphytum bracteosum were better, and the result of Dieffenbachia picta obtained by using the proposed method was worse compared with that of the other two plants because of its morphology characteristics.

The numbers of points resulting from the classification of the three plants by using the three methods are listed in Table 6.

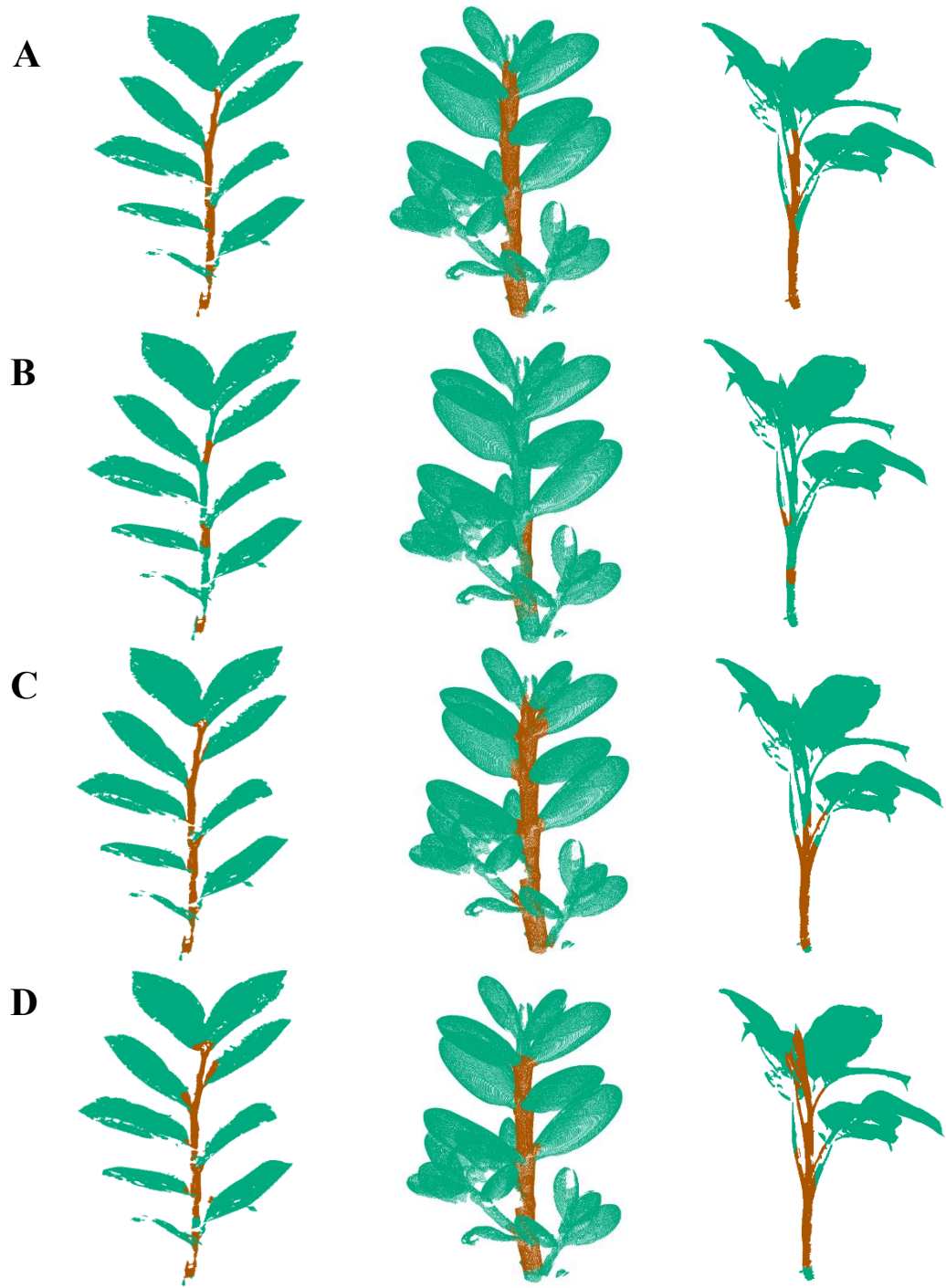

FIG. 12. Classification results of standard classification (A), random selection method (B), artificial selection method (C) and automated classification method (D) of Zamioculas zamiifolia, Pachyphytum bracteosum and Dieffenbachia picta (from left to right).

| Plant | Number of leaf points | Number of stem points |
|---|---|---|
| Zamioculas zamiifolia | 982236 | 61984 |
| Pachyphytum bracteosum | 789775 | 81802 |
| Dieffenbachia picta | 2601906 | 89625 |

TABLE 5. *Numbers of points resulting from standard classification.*

| Plant | Classification Method | | | | | |
|---|---|---|---|---|---|---|
| | Random selection method | | Artificial selection method | | Automated selection method | |
| | Number of leaf points | Number of stem points | Number of leaf points | Number of stem points | Number of leaf points | Number of stem points |
| Zamioculas zamiifolia | 1025454 | 18766 | 973135 | 71085 | 958139 | 86081 |
| Pachyphytum bracteosum | 852936 | 18641 | 757831 | 113746 | 791318 | 80259 |
| Dieffenbachia picta | 2678107 | 13424 | 2614521 | 77010 | 2428212 | 263319 |

TABLE 6. *Numbers of points resulting from classification of the three plants by using the three methods.*

Because of the large difference in the numbers of leaf points and stem points, to evaluate the classification results of the three different methods more reasonably, the confusion matrix and kappa coefficient were calculated for evaluation.

The following shows the specific accuracy analysis of the classification results obtained by using the three methods.

First, several evaluation indicators were set to compare the different results in detail:

(1) Standard leaf point set: $L_s$.

(2) Classified leaf point set: $L_c$.

(3) Standard stem point set: $S_s$.

(4) Classified stem point set: $S_c$.

(5) Cardinal number of set $A$ (that is, the number of elements of set $A$): $card(A)$.

Second, based on the definitions mentioned above and the related definitions of the confusion matrix, the following additional indicators can be set:

True Positives $TP$, which denotes the number of correctly classified leaf points.

True Negatives $TN$, which denotes the number of correctly classified stem points.

False Positives $FP$, which denotes the number of mistakenly classified leaf points.

False Negatives $FN$, which denotes the number of mistakenly classified stem points.

These indicators can be calculated using the following equations:

$$TP = card(L_S \cap L_C) \tag{1}$$

$$TN = card(S_S \cap S_C) \tag{2}$$

$$FP = card(L_C) - TP \tag{3}$$

$$FN = card(S_C) - TN \tag{4}$$

Then, the confusion matrices of the three plant classification results obtained by using the three methods can be calculated, which are shown as Fig. 13, in which A, B, and C denote the random selection method, artificial selection method, and automated classification method, respectively, and D, E, and F denote Zamioculas zamiifolia, Pachyphytum bracteosum, and Dieffenbachia picta, respectively.

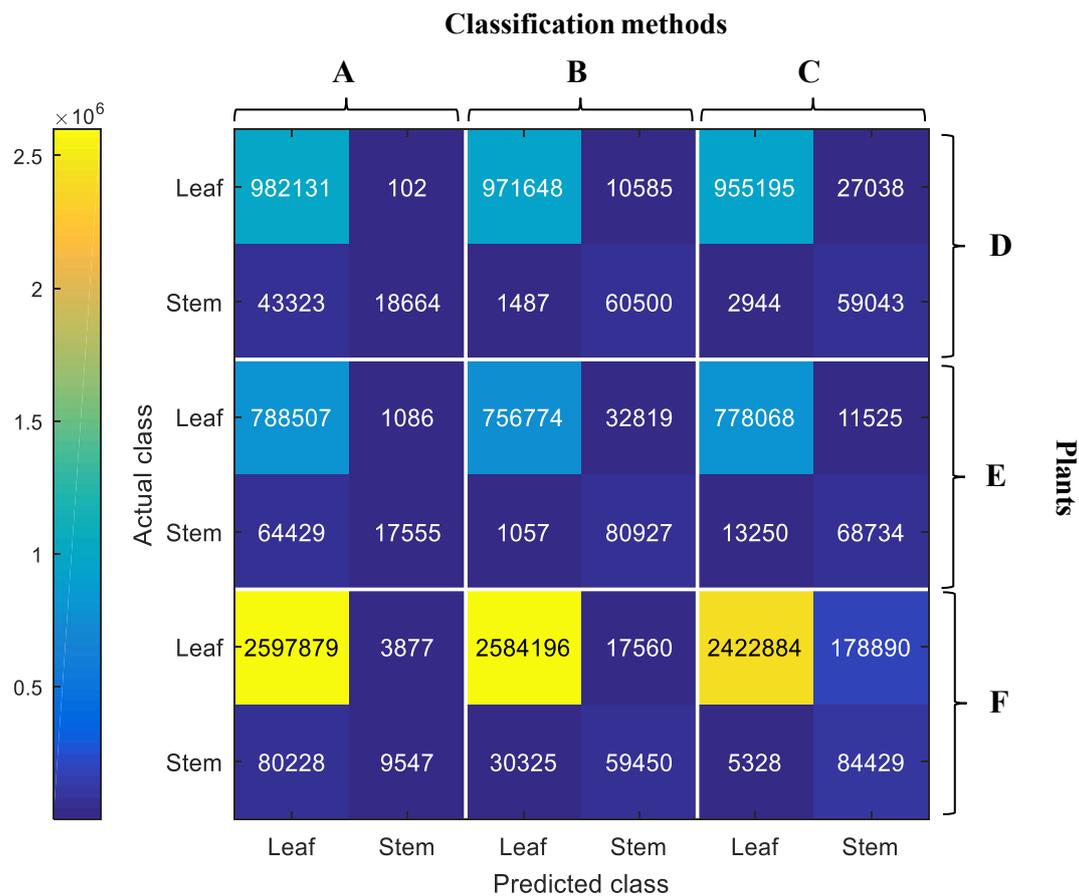

FIG. 13. Confusion matrices of the experiments.
Column: A. Random selection method. B. Artificial selection method.
C. Automated classification method.
Row: D. Zamioculas zamiifolia. E. Pachyphytum bracteosum. F. Dieffenbachia picta.

Based on the confusion matrices above, the kappa coefficients of the three methods were calculated and are listed in Table 7.

| Plant | Kappa coefficient | | |
|---|---|---|---|
| | Random selection method | Artificial selection method | Automated selection method |
| **Zamioculas zamiifolia** | 0.4470 | 0.9031 | 0.7825 |
| **Pachyphytum bracteosum** | 0.3254 | 0.8075 | 0.8316 |
| **Dieffenbachia picta** | 0.1779 | 0.7038 | 0.4509 |

TABLE 7. *Kappa coefficients of the three methods.*

## 4. Discussion

According to the classification results (Fig. 12) and the kappa coefficients (Table 7), the proposed method has some advantages in the classification experiments. In the classification of Zamioculas zamiifolia and Pachyphytum bracteosum, the proposed method and the artificial selection method had very similar performances. When the point cloud data of Zamioculas zamiifolia were used in the experiments, the artificial selection method yielded the best result, and the difference in the kappa coefficients between the artificial selection method and automated selection was 0.1206, which was very close. The proposed method had a kappa coefficient of 0.8316, which was the best result among the three methods when the point cloud data of Pachyphytum bracteosum was processed, and the kappa coefficients of the proposed method and the artificial selection method were 0.8316 and 0.8075, respectively, which were also very close. However, when the point cloud data of Dieffenbachia picta were processed, the kappa coefficient of the artificial selection method yielded the best result among the three methods, which were 0.1779, 0.7038 and 0.4509. In terms of efficiency, the random method is the fastest but inaccurate. The artificial selection method is time-consuming because of the need to repeatedly select leaf sample points and stem sample points for a better classification result. However, the proposed method is automated and more efficient, with a good performance.

Yet, the proposed method still has some disadvantages. First, it has the requirement for the integrity of the data, that is, the robustness of the algorithm still needs to be improved. For example, the Pachyphytum bracteosum stem is thick, which accounts for a larger proportion of the total volume. When the stem sample points were selected, if the entire stem was not completely covered by sample points, then some of the stem points would not be classified correctly. Second, some leaves with special shapes will interfere with the algorithm and affect the classification accuracy. For example, a slender leaf in the middle part of Dieffenbachia picta is shaped like a stem, which was classified as a stem point (Fig. 14); thus, the kappa coefficient of the automated classification method had a worse result compared to those of the other two plants.

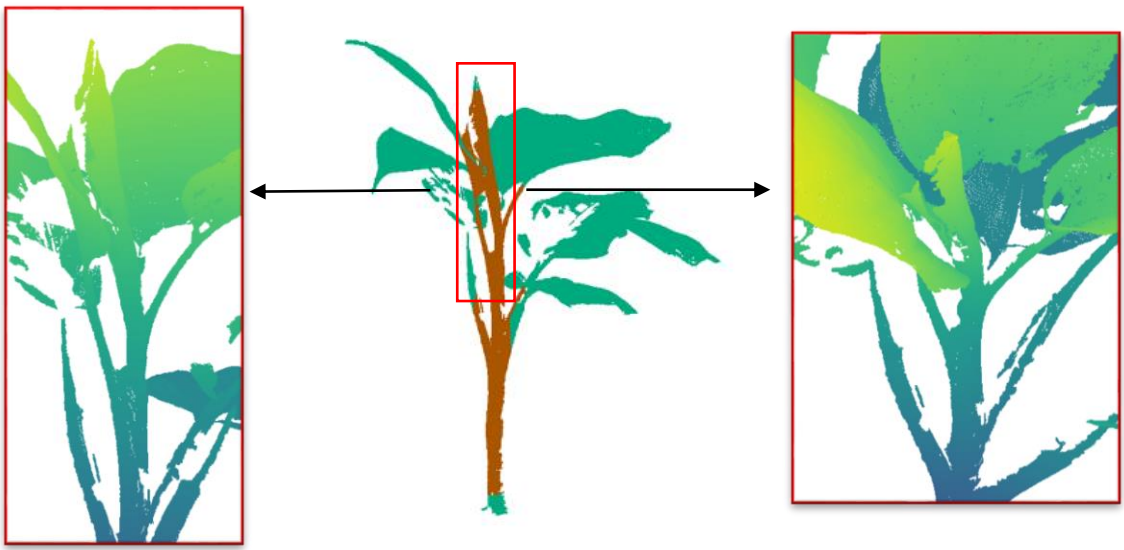

FIG. 14. The suspicious part of Dieffenbachia picta.

The slender leaf in the middle part of Dieffenbachia picta was deleted, and the processed data were tested by using the three methods. The new classification results and the new kappa coefficients are shown and listed in Fig. 15 and Table 8. Obviously, after deleting the slender leaf, the kappa coefficient of the automated classification result was improved, and the results of the artificial selection method and automated classification method became more similar.

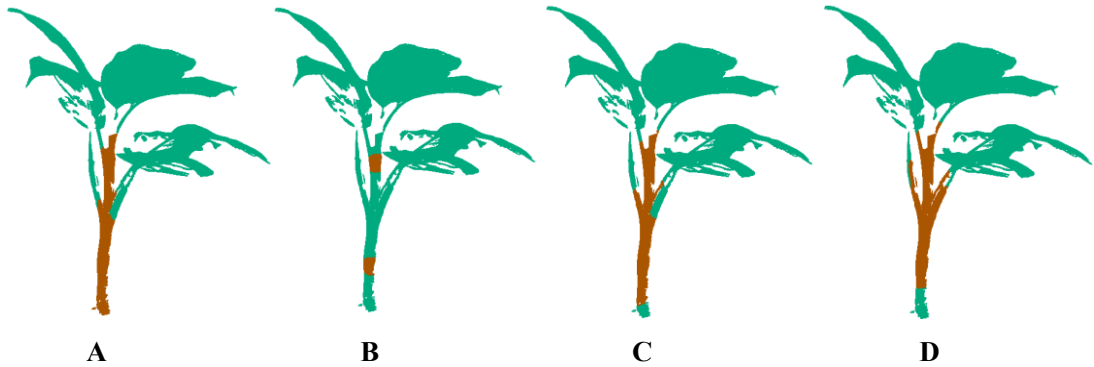

|   |   |   |   |
|---|---|---|---|
| A | B | C | D |

FIG. 15. New classification results of Dieffenbachia picta obtained by using the three methods. A. Standard classification results. B. The result of the random selection method. C. The result of the artificial selection method. D. The result of automated classification.

| Plant | Kappa coefficient | | |
|---|---|---|---|
| | Random selection method | Artificial selection method | Automated selection method |
| Dieffenbachia picta | 0.3348 | 0.9210 | 0.7851 |

TABLE 8. *Kappa coefficients of the three methods with the processed Dieffenbachia picta data.*

Some similar studies have reported their accuracies of classification in publications. Ferrara et al. (2018) firstly used the DBSCAN method to classify wood and non-wood points in point cloud data obtained with a terrestrial laser scanning (TLS) instrument. In his experiment, 7 randomly selected cork oak trees were scanned from four different directions to have a complete reconstruction of the whole canopy and branching structure as well as to mitigate shadowing effects. Then a voxel grid filtering approach and three different thresholds were used to help the DBSCAN method classify the tree points into wood points and non-wood points. The reported Kappa coefficients of the method with different thresholds is from 0.75 to 0.88.

Tao et al. (2015) also scanned two real trees and simulated a virtual tree to carry out the classification. These trees were simplified to skeleton points consisting of circle centers and line nodes. And wood points and leaf points were separated by using the Dijkstra's shortest-path algorithm and the Kd-Tree method. The Kappa coefficients of this geometric method is

from 0.79 to 0.89.

Obviously, the Kappa coefficients in our study are higher than the above two reported experiments. First reason maybe the automated samples selection in our study which reduces the error of manual selection. Second, the small sizes of the potted plants result in less occlusions and shades between leaves and stems which is favorable to classification. Third, the scanner used in our experiment is a kind of desktop scanner which has a better accuracy than the TLS instruments. More accurate acquisition of point cloud data means less noise points and more guarantees on accuracy.

In general, the results show that the automated classification method can effectively classify the stems and leaves of plants with good shapes; it has a more accurate classification result compared to that of the random selection method and less time cost compared to that of the artificial selection method. Furthermore, the proposed method can avoid subjective influence and misoperation when selecting sample points. The experimental results also show its applicability and the potential for improvement.

## 5. Conclusions

The point cloud data of plants provide a nondestructive solution for collecting, monitoring and analyzing the status of plants and their organs. This study proposed a feasible and automated method for classifying the point cloud data of plants into two classes: the leaf class and the stem class. The training sample sets of the two classes can be automatically identified by using the convex hull and the 2D distribution of the point cloud data. The accuracy of three classification methods was evaluated, and the results show that the proposed method has the advantage in terms of the classification accuracy. The proposed automated method has the potential to improve the efficiency of research on monitoring or analyzing plants. Future work could focus on improving the classification accuracy and the robustness via application to different plants.

## Acknowledgements

This work was supported by the Fundamental Research Funds for the Central Universities (No. 2015ZCQ-LY-02); the Fundamental Research Funds for the Central Universities (BLX201928); the State Scholarship Fund from China Scholarship Council (CSC No. 201806515050).